\def\year{2021}\relax
\documentclass[letterpaper]{article} 
\usepackage{aaai21}  
\usepackage{times}  
\usepackage{helvet} 
\usepackage{courier}  
\usepackage[hyphens]{url}  
\usepackage{graphicx} 
\usepackage{natbib}  
\usepackage{caption} 
\usepackage{bm}
\usepackage{booktabs}
\usepackage{array}
\usepackage{multirow}
\usepackage{makecell}
\usepackage{arydshln}
\usepackage{graphicx}
\usepackage{color}
\usepackage{amsmath}
\usepackage{amssymb}
\usepackage{tikz}
\newcommand*{\Scale}[2][4]{\scalebox{#1}{$#2$}}  

\title{Few-shot Learning for Multi-label Intent Detection}

\author{
	Yutai Hou,
	Yongkui Lai,
	Yushan Wu,
	Wanxiang Che\thanks{$\ $ Corresponding author.}~,
	Ting Liu
	\\
	\normalfont{School of Computer Science and Technology, Harbin Institute of Technology, China} \\
	{\tt \{ythou, yklai, car, tliu\}@ir.hit.edu.cn,} {\tt wuyushan@hit.edu.cn} \\
}

\begin{document}

\maketitle

\begin{abstract}
In this paper, we study the few-shot multi-label classification for user intent detection. 
For multi-label intent detection, state-of-the-art work estimates \textit{label-instance relevance scores} and uses a \textit{threshold} to select multiple associated intent labels. 
To determine appropriate \textit{thresholds} with only a few examples, we first learn universal thresholding experience on data-rich domains, and then adapt the thresholds to certain few-shot domains with a calibration based on non-parametric learning.
For better calculation of \textit{label-instance relevance score}, we introduce label name embedding as anchor points in representation space, which refines representations of different classes to be well-separated from each other.
Experiments on two datasets show that the proposed model significantly outperforms strong baselines in both one-shot and five-shot settings.\footnote{Data and code are available at \url{https://github.com/AtmaHou/FewShotMultiLabel}}

\end{abstract}

\section{Introduction}

\begin{figure}[t]
	\centering
	\begin{tikzpicture}
	\draw (0,0 ) node[inner sep=0] {\includegraphics[width=1\columnwidth, trim={9.55cm 4.8cm 5.5cm 6.0cm}, clip]{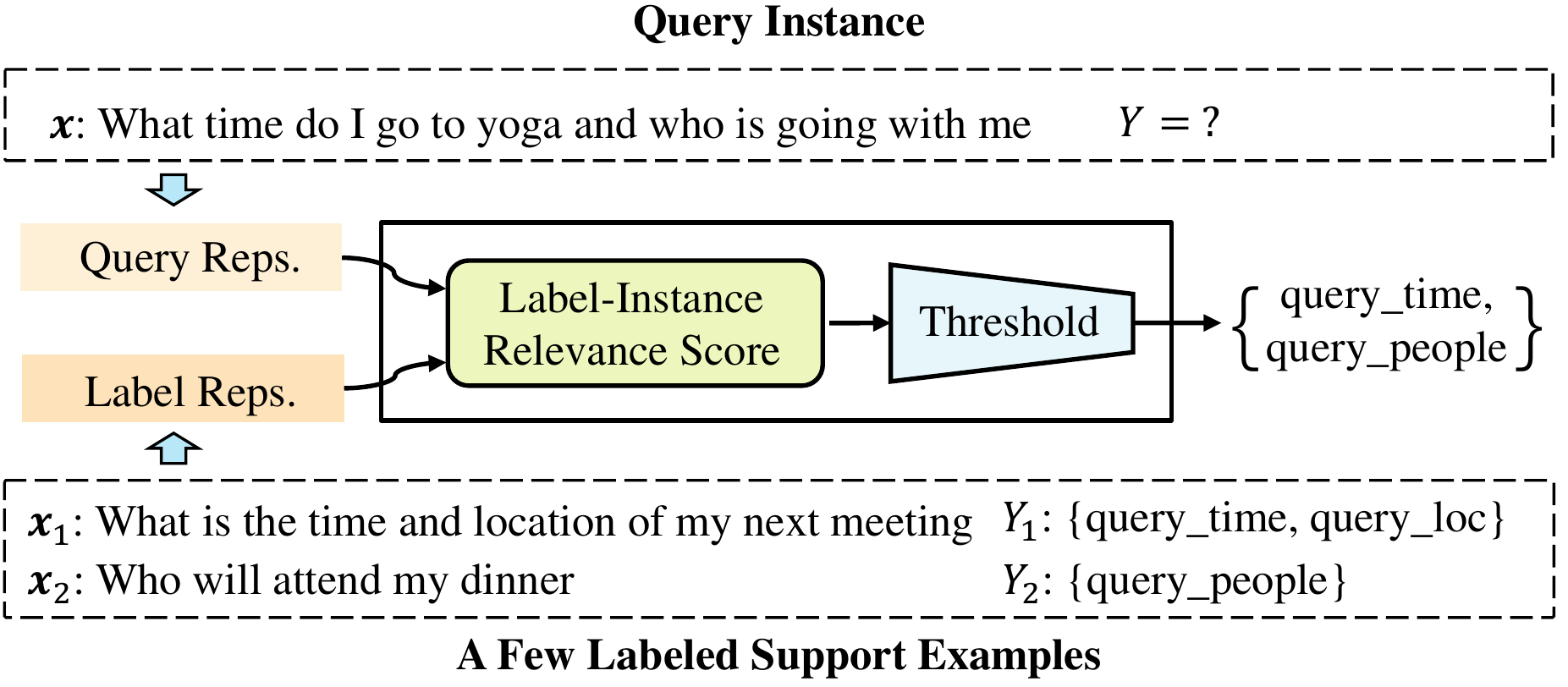}};
	\end{tikzpicture}  
	\caption{\footnotesize
		Example for one-shot multi-label intent detection.
	}\label{fig:intro1}
\end{figure}

Intent detection, a fundamental component of task-oriented dialogue system \cite{young2013pomdp}, is increasingly raising attention as a Multi-Label Classification (MLC) problem \cite{xu2013exploiting,qin2020td}, since a single utterance often carries multiple user intents (See examples in Fig \ref{fig:intro1}). 
In real-world scenarios, intent detection often suffers from lack of training data, because dialogue tasks/domains change rapidly and new domains usually contain only a few data examples. 
Recent success of Few-Shot Learning (FSL) presents a promising solution for such data scarcity challenges. 
It provides a more human-like learning paradigm that generalizes from only a few learning examples (usually one or two per class) by exploiting prior experience.

State-of-the-art works for multi-label intent detection focus on threshold-based strategy, where a common practice is estimating \textit{label-instance relevance scores} and picking the intent labels with score higher than a \textit{threshold} value \cite{xu2017convolutional,gangadharaiah2019joint,qin2020td}. 
Usually, the coordination and respective quality of the two modules, i.e. thresholding and relevance scoring, are crucial to the performance of MLC models. 
However, in few-shot scenarios, such multi-label setting poses unique challenges for both threshold estimation and label-instance relevance scoring.

For thresholding, previous works explore to tune a fixed threshold \cite{gangadharaiah2019joint,qin2020td} or to learn thresholds from data \cite{xu2017convolutional}.
But, these thresholds work well only when learning examples are sufficient. 
In few-shot scenarios, it is pretty hard to determine appropriate thresholds with only a few examples.
Besides, it is also difficult to directly transfer the pre-learned thresholds due to the domain differences, such as differences in label number per instance, score density and scale.

Estimation of the label-instance relevance scores is also challenging.
Few-shot learning has achieved impressive progress with similarity-based methods  \citep{matching,DBLP:conf/iclr/BaoWCB20}, where the relevance scores can be modeled as label-instance similarities. 
And the label representations can be obtained from corresponding support examples. 
Unfortunately, despite huge success in previous single-label tasks, these similarity-based methods become impractical for multi-label problems. 
When instances have multiple labels, representations of different labels may be obtained from the same support examples and become confused with each other.
For the example in Fig \ref{fig:intro1}, intents of \textit{query\_time} and \textit{query\_loc} share the same support example $\bm{x}_1$ and thus have the same label representation, 
which makes it impossible to predict correct labels with similarity scores. 

In this paper, we study the few-shot learning problem of multi-label intent detection and propose a novel framework to tackle the challenges from both thresholding and label-instance relevance scoring.

To solve the thresholding difficulties of prior-knowledge transferring and domain adaption with limited examples, we propose a \textit{Meta Calibrated Threshold} (MCT) mechanism that first learns universal thresholding experience on data-rich domains, then adapts the thresholds to certain few-shot domains with a Kernel Regression based calibration. 
Such combination of universal training and domain-specific calibration allows to estimate threshold using both prior domain experience and new domain knowledge. 

To tackle the challenge of confused label representation in relevance scoring, we propose the \textit{Anchored Label Representation} (ALR) to obtain well-separated label representations.
Inspired by the idea of embedding label name as anchor points to refine representation space \cite{anchor}, ALR uses the embeddings of label names as additional anchors and represents each label with both support examples and corresponding anchors. 
Different from the previous single-label intent detection that uses label embedding as additional features \cite{chen2016zero}, our label embeddings here have unique effects of separating different labels in metric space.

Finally, to encourage better coordination between thresholding and label-instance relevance scoring, we introduce the Logit-adapting mechanism to MCT that automatically adapts thresholds to different score densities. 

Experiments on two datasets show that our methods significantly outperform strong baselines. 
Our contributions are summarized as follows: 
(1) We explore the few-shot multi-label problem in intent detection of task-oriented dialogue, which is also an early attempt for the few-shot multi-label classification. 
(2) We propose a Meta Calibrated Threshold mechanism with Kernel Regression and Logits Adapting that estimates threshold using both prior domain experience and new domain knowledge.
(3) We introduce the Anchored Label Representation to obtain well-separated label representation for better label-instance relevance scoring.

\section{Notions and Preliminaries}
To ease understanding, we briefly introduce the task of multi-label classification and few-shot learning here. 

\subsection{Multi-label Classification}\label{sec:mlc}
Multi-label task studies the classification problem where each single instance is sociated with a set of labels simultaneously. 
Suppose $\mathcal{X}$ denotes instance space and $\mathcal{Y}=\{y_1, y_2, \ldots, y_N \}$ denotes label space with $N$ possible labels.
Multi-label task learns a function $h(\cdot): \mathcal{X} \rightarrow 2^{\mathcal{Y}}$ from multi-label training data $T=\{(\bm{x}_i, Y_i)\}_i^{N_T}$, 
where $N_T$ is the size of datasets.
For each learning example $(\bm{x}_i, Y_i)$, 
$\bm{x}_i \in \mathcal{X}$ is $l$-dimensional input and
$Y_i \subseteq \mathcal{Y}$ is the corresponding label set.  
Then for an unseen instance $\bm{x}$, the classifier predicts $Y=h(\bm{x}) \subseteq \mathcal{Y}$ as the associated label set. 

In most cases \cite{zhang2013review}, 
multi-label model learns a real-value function $f: \mathcal{X} \rightarrow \mathcal{Y}$. 
$f(\bm{x}, y)$ evaluates the \textit{label-instance relevance score}, which reflects the confidence of label $y \in \mathcal{Y}$ being the proper label of $x$. 
Then multi-label classifier is derived as $h(\bm{x}) = \{y \mid f(\bm{x}, y) > t, y \in \mathcal{Y}\}$, 
where $t$ is the \textit{threshold} value. 

\subsection{Few-shot Learning}
Few-shot learning extracts prior experience that allows quick adaption on new tasks \cite{finn2018learning}. 
The learned prior experience often includes meta knowledge general to different domains and tasks, such as similarity metric and model architecture.
On the new task, few-shot model uses these prior knowledge and a few labeled examples (\textit{support set}) to predict the class of an unseen item (\textit{query}).

Few-shot learning is often achieved with similarity based methods, 
where models are usually first trained on a set of source domains (tasks) $\left\{\mathcal{D}_1, \mathcal{D}_2, \ldots \right\}$, 
then directly work on another set of unseen target domains (tasks) $\left\{\mathcal{D}_1', \mathcal{D}_2', \ldots \right\}$ without fine-tuning. 
On each target domain, given a query $\bm{x}$,  
model predicts the corresponding label $y$ by observing a labeled support set $S = \left\{(\bm{x}_i,y_i)\right\}_{i=1}^{N_S}$.
$S$ usually includes $k$ examples (K-shot) for each of $N$ labels (N-way).

For few-shot multi-label intent detection, we define each query instance as user utterance with a sequence of words $\bm{x} = (x_1, x_2, \ldots, x_l)$. And instead of predicting single label, model predicts a set of intent labels $Y = \{y_1, y_2, \ldots, y_m \}$ with multi-label support set $S = \left\{(\bm{x}_i,Y_i)\right\}_{i=1}^{N_S}$.



\section{Method}
\begin{figure}[t]
	\centering
	\begin{tikzpicture}
	\draw (0,0 ) node[inner sep=0] {\includegraphics[width=1\columnwidth, trim={9.55cm 0.7cm 11.5cm 1.3cm}, clip]{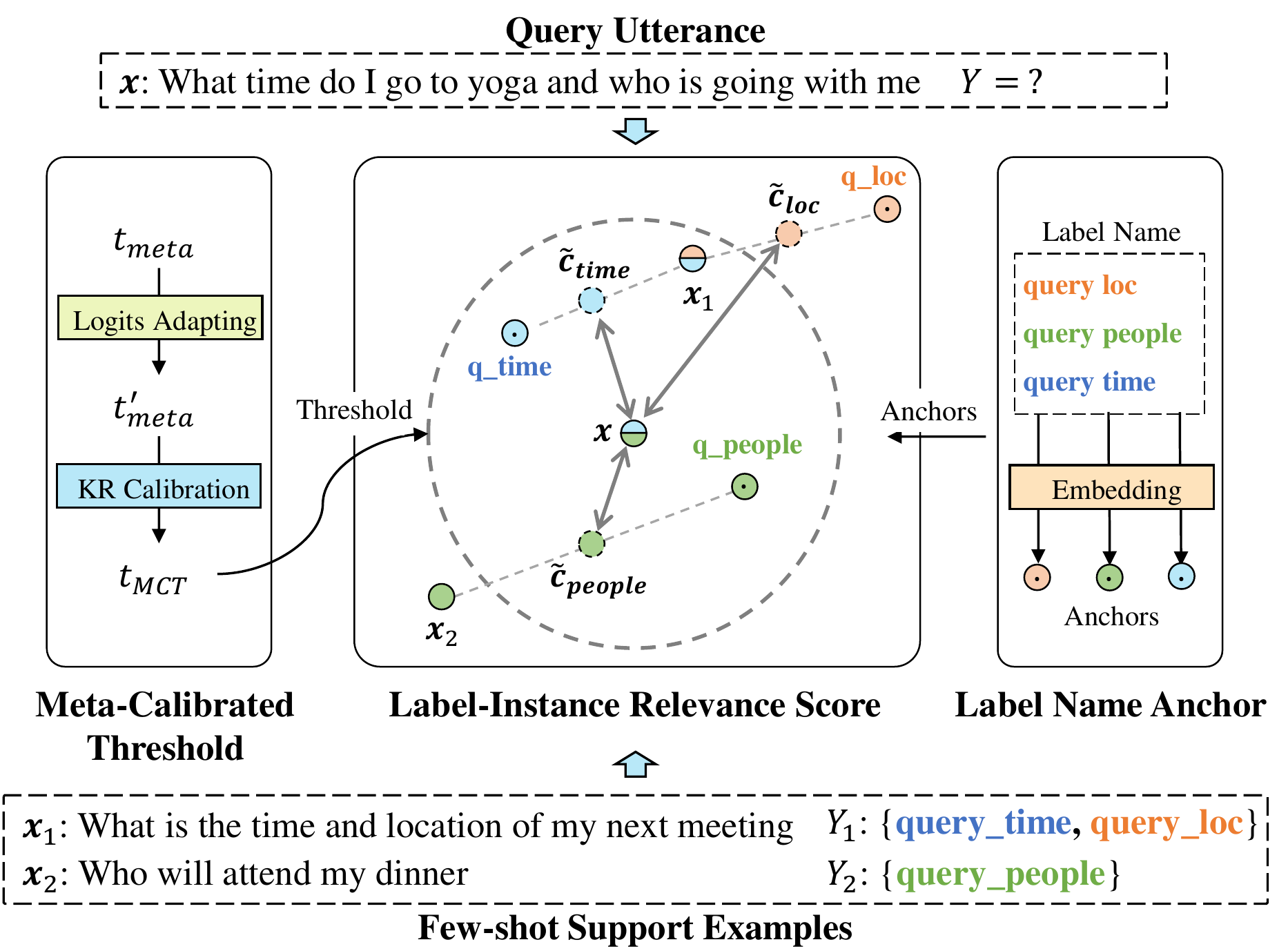}};
	\end{tikzpicture}  
	\caption{\footnotesize
		Proposed framework for few-shot multi-label intent detection. 
		For each query $\bm{x}$,  
		we compute label-instance relevance scores according to its similarity to each Anchored Label Representation $\tilde{\bm{c}}$.
		Then we pick the labels that have score higher than a threshold value $t$ derived from the proposed Meta-Calibrated Threshold mechanism. 
	}\label{fig:overall}
\vspace*{-3mm}
\end{figure}

In this section, we first present the overview of our framework, and then introduce the proposed Meta Calibrated Threshold and Anchored Label Representation.
In the last, we introduced how the entire framework is optimized.

The workflow of our framework is shown in Fig \ref{fig:overall}: given an unseen query utterance and a labeled support set, we first gather representation of each label from support set, and calculate label-instance relevance scores with sentence-label similarity. Then we use a threshold to dichotomize the label space into relevant and irrelevant label sets.
In the framework, the MCT helps to estimate reasonable thresholds under few-shot scenarios, and the ALR provides well-separated label representation.


\subsection{Proposed Framework for Few-shot Multi-label Intent Detection}\label{sec:fsmlc}

We define the few-shot multi-label classification (MLC) similar to the aforementioned normal MLC.
Given a query sentence $\bm{x}$ and a support set $S$, our framework predicts the associated label set $Y$ as:
\[
\Scale[0.85]{
	Y= h(\bm{x}, S) = \{y \mid f(\bm{x}, y, S) > t, y \in \mathcal{Y}\},
}
\]
where $f$ calculates label-instance relevance scores, and $t$ is threshold value.

To achieve few-shot multi-label classification, we adopt the prevailing similarity-based method to calculate the label-instance relevance scores.

Firstly, we derive the representation of each label from the support set $S$. 
Supposing $\bm{c}_{i}$ is the representation vector of label $y_i$, we compute relevance score between query sentence $\bm{x}$ and label $y_i$ as follow:
\[
\Scale[0.85]{
	f(\bm{x}, y_i, S) = \textsc{Sim}(E(\bm{x}), \bm{c}_i),
}
\]
where $E(\cdot)$ is an embedder function. 
$\textsc{Sim}$ is a similarity function, and we use the dot-product similarity. 
We adopt BERT \cite{BERT} as the embedder, and the sentence embedding $E(\bm{x})$ is calculated as the averaged embedding of its tokens.
To get well-separated label representations, we adopt the \textbf{Anchored Label Representation} to obtain $\bm{c}_i$.

Then, we estimate a threshold value $t$ that integrates both prior knowledge from source domains and observation of examples from target domains. 
To achieve this, we propose the \textbf{Meta Calibrated Threshold} to estimate threshold $t$.

\subsection{Meta Calibrated Threshold}\label{sec:threshold}
In this section, we introduce a thresholding method for few-shot learning setting. 
In few-shot learning setting, models are trained and tested on different domains, which often have different preferences for threshold selection. 
Further, it is necessary to label each instance with different thresholds, because instances vary in label number and density of label-instance relevance scores.

To achieve this, we first learn a domain-general meta threshold, and then calibrate it to adapt to both target domain and specific queries.

%

\paragraph{Meta Threshold with Logits-Adaptiveness}
To achieve domain-general thresholding, we present a Meta Threshold $t_\text{meta}$ that has automatic adaptability and is jointly optimized on various domains.

\begin{figure}[t]
	\centering
	\begin{tikzpicture}
	\draw (0,0 ) node[inner sep=0] {\includegraphics[width=1\columnwidth, trim={0cm 13cm 20cm 0cm}, clip]{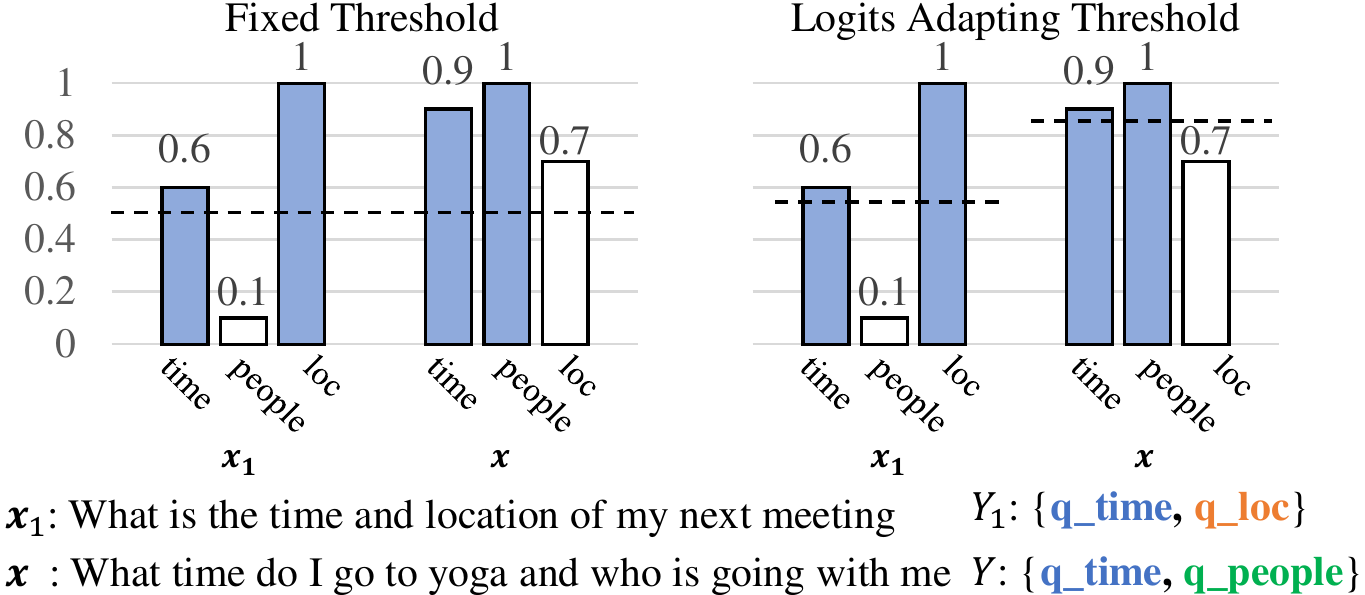}};
	\end{tikzpicture}  
	\caption{\footnotesize
		Example of fixed threshold and logit-adapting threshold.
		Colored score-bars correspond to correct labels. 
		It is impossible to find a fixed threshold fitting both $\bm{x}_1$ and $\bm{x}$, but logit-adapting threshold can adapt to both cases with $r$ = 0.5.
	}\label{fig:ea}
\end{figure}

For automatic adaptability, we propose the Logit-Adaptive mechanism for meta threshold, which automatically adapts the threshold to both specific queries and domains.
Specifically, considering that instances vary in scale/density of relevance scores and the value of threshold is always between the maximum and the minimum score, we propose to set the threshold as an interpolation of maximum and minimum scores:
\[
\Scale[0.85]{
	t_\text{meta}=T(\bm{x}, S)=\displaystyle  r \max_{\forall y \in \mathcal{Y}} f(\bm{x}, y, S)  + (1- r) \min_{\forall y \in \mathcal{Y}} f(\bm{x}, y, S)\label{eq:iat},
}
\]
where $T(\cdot)$ is the thresholding function and $r$ is the interpolation rate learned in source domains.\footnote{Extreme cases of outputting all labels can be covered by imposing a small negative perturbation on $t_\text{meta}$.}
As shown in Fig \ref{fig:ea}, such interpolation-based thresholds vary for different queries by adapting to different densities of label-instance relevance scores, and are more general than fixed thresholds.

Besides, Logit-Adaptive mechanism also promotes better coordination between thresholding and relevance scoring.

\paragraph{Threshold Calibration with Kernel Regression}
Till now, we learn a Meta Threshold $t_\text{meta}$, which is general to various domains but lacks domain-specific knowledge. 
To remedy this, we estimate a domain/query-specific threshold $t_\text{est}$ by observing the support set, and use $t_\text{est}$ to calibrate the Meta Threshold.
However, owing to the absence of golden thresholds, it is hard to directly learn a model to estimate thresholds. 
Therefore, we turn to estimate the number of labels and indirectly deduce the threshold. 

To estimate label numbers with domain-specific knowledge, we adopt \textit{Kernel Regression} (KR) \cite{kernelregression} and estimate the label number of a query according to its similarity to support examples.
As a non-parametric method, KR can work on unseen domains without tuning.
Compared to other non-parametric regression methods, such as KNN regression, KR allows to make use of all support examples and consider the distance influences. 

Formally, given a support set $S$, we estimate the label number $n$ of query $\bm{x}$ as the weighted average label number of support examples, where the weights are computed as kernel similarity between query and support examples: 
\[
\Scale[0.85]{
	n = \displaystyle \frac{1}{Z} \sum_{\forall (\bm{x'}, \bm{y'}) \in S} {{\rm Kernel}(\tilde{E}(\bm{x}), \tilde{E}(\bm{x'});\lambda) \cdot |\bm{y'}| }.
}
\]
Here, $Z$ is the normalizing factor, and we use Gaussian Kernel: ${\rm Kernel}(\bm{a}, \bm{b};\lambda)=\exp(-(\bm{a}-\bm{b})^2/\lambda)$, where $\lambda$ is bandwidth factor.
$\tilde{E}(\bm{x})$ is a feature extractor that returns a feature vector related to the label number of a sentence $\bm{x}$.
For the intent detection task of this paper, we consider the linguistics features related to the intent number of a sentence, including  
\textit{sentence length}, \textit{\# of conjunctions}, \textit{\# of predicates}, \textit{\# of punctuations}, \textit{\# of interrogative pronouns} and encode these features with an MLP projection layer.\footnote{We extract these linguistics features with StanfordTagger-Base \cite{toutanova2003feature}.}

Then, we derive a domain/query-specific threshold $t_\text{est}$ from the estimated label number $n$. 
Specifically, we find a threshold value $t_\text{est}$ that filters out top-$n$ label-instance relevance scores of $\bm{x}$.
One intuitive idea is to directly use the $(n+1)_\text{th}$ largest score as threshold.
But, such threshold is derived from only one label-instance relevance score.
So we further improve it to make use of all relevance scores by directly estimating threshold with the learned kernel weights:
\[
\Scale[0.85]{
	t_\text{est} = \displaystyle \frac{1}{Z} \sum_{\forall (\bm{x'}, \bm{y'}) \in S} {{\rm Kernel}(\tilde{E}(\bm{x}), \tilde{E}(\bm{x'});\lambda) \cdot T'(|\bm{y'}|; \bm{x}, S, f) }, 
}
\]
where $T'(n; \bm{x}, S, f)$ is a function that returns the $(n+1)_\text{th}$ largest label-instance relevance scores of query $\bm{x}$.

Finally, we use query-specific threshold $t_\text{est}$ to calibrate the domain-general meta threshold $t_\text{meta}$. 
The final threshold for query $\bm{x}$ is computed as:
\begin{equation}
\Scale[0.85]{
	t = \alpha \times t_\text{meta} + (1 - \alpha) \times t_\text{est},\label{eq:alpha}
}
\end{equation}
where $\alpha$ is hyper-parameter that measures the importance of the prior thresholding experience.

\subsection{Anchored Label Representation}\label{sec:label_reps}
Label representation is essential in the label-instance relevance score. 
To get high-quality label-instance similarity modeling, label representations should be 
(1) well-separated from each other and
(2) able to fully express the semantic information of the corresponding category.

\paragraph{Label Representation for Few-shot Learning}
For few-shot learning, the label representations are mainly obtained from support set examples. 
One of the most classic ideas is to get a prototypical representation of each category as label representation \cite{prototypical}.
And the prototypical representation of label $y_i$ is calculated as the averaged embedding of support examples: $\bm{c}_i = \frac{1}{M_i}\sum_j^{M_i}{ E(\bm{x}_j)}$,
where each $\bm{x_j} \in \{\bm{x} \mid (\bm{x}, Y) \in S \wedge y_i \in Y \}$ is the support instance labeled with $y_i$, and $M_i$ is the total number of such instances in support set $S$. 

Such label representations have no constraint on the separability between labels. 
Further, in multi-label setting, different labels may share the same support examples.  
This can lead to mix-up and ambiguity between label representations.

\paragraph{Represent Label with Anchor} 
Because obtaining the label representation with only support examples leads to ambiguity, 
we propose to additionally represent the labels with label-specific anchors, 
which emphasizes the difference between different categories. 
Label names are often naturally well separated from each other and contain good expressions of category-specific semantics \cite{anchor}.
Therefore, intuitively, we use semantic embedding of label names as the anchors and represent each label with both anchor and support examples.
For label $y_i$, we compute the anchored label representation $\tilde{\bm{c}}_i$ with an interpolation factor $\beta$:
\begin{equation}
\Scale[0.85]{
	\tilde{\bm{c}}_i = \beta \times E(y_i) + (1 - \beta) \times \bm{c}_i,\label{eq:beta}
}
\end{equation}
where $\bm{c}_i$ is the prototypical representation obtained with support examples. 
Here, label name embedding $E(y_i)$ acts as a deflection of the prototypical representation vector. 
This allows label representations to be separated from each other and better describe the category semantics.

\subsection{Optimization}
Following \citet{matching}, we train the MLC framework with a series of few-shot learning episodes, where each episode contains a few-shot support-set and a query-set. 
Simulating few-shot situation on data rich domains ensures consistence between training and few-shot testing. 
Besides, the framework is optimized on different domains alternatively, which encourages both meta threshold $t_\text{meta}$ and label-instance relevance scoring function $f$ to be domain-general. 
We take the Sigmoid Cross Entropy loss \cite{cui2019class} for MLC training:
\[
\Scale[0.85]{
	\text{Loss} =\\ 
	\frac{1}{N} \sum_i^N \Big \{ \mathbb I(y_i \notin Y^*) \cdot \sigma(f_{y_i}) - \mathbb I(y_i \in Y^*) \cdot \sigma(f_{y_i}) \Big \},
}
\]
where $N$ is the number of possible labels and $f_{y_i}=f(\bm{x}, y_i, S)$.
$Y^*$ is the golden label set. 
$ \mathbb I(\cdot)$ is an indicator function.\footnote{~$\mathbb I(True)=1$ and $\mathbb I(False)=0$} $\sigma$ is the Sigmoid function.
Due to the undifferentiable process of picking thresholds with label numbers, we pre-train the kernel parameters before the whole framework learning process, i.e. bandwidth and MLP projection layer, on source domains with the loss of $\text{MLE}(n_\text{est}, n_\text{gold})$.

This is a meta-learning process that learns meta parameters (of Meta Threshold, Kernel Regression, similarity metric computing) to improve non-parametric learning of unseen few-shot tasks, a.k.a learning to learn.


\section{Experiment}
We evaluate our method on the multi-label intent detection task of 1-shot/5-shot setting, 
which transfers knowledge from source domains (training) to an unseen target domain (testing) containing only a 1-shot/5-shot support set.

\paragraph{Dataset} 

We conduct experiments on public dataset TourSG \cite{toursg} and introduce a new multi-intent dataset StanfordLU.
These two datasets contain multiple domains and thus allow to simulate the few-shot situation on unseen domains.
TourSG (DSTC-4) contains 25,751 utterances annotated with multiple dialogue acts and 6 separated domains about touristic information for Singapore:
Itinerary (It),  Accommodation (Ac), Attraction (At), Food (Fo), Transportation (Tr), Shopping (Sh).
StanfordLU is an re-annotated version of Stanford dialogue dataset \cite{eric2017key} containing 8,038 user utterances from 3 domains:
Schedule (Sc), Navigate (Na), Weather (We). We re-annotate each utterance with intent labels, which are not included in the original dataset.



\begin{table}[t]
	\centering
	\footnotesize
	\begin{tabular}{rrrrr}
		\toprule
		\multicolumn{1}{c}{\textbf{Domain}} & \multicolumn{1}{c}{\textbf{1-shot $\bm{|S|}$}} & \multicolumn{1}{c}{\textbf{5-shot$\bm{|S|}$}} & \multicolumn{1}{c}{\textbf{P. ML}} & \multicolumn{1}{c}{\textbf{$|\mathcal{Y}|$}} \\
		\midrule
		{\textbf{It}} & 12.56 & 48.44 & 22.7\% & 16 \\
		{\textbf{Ac}} & 13.93 & 59.95 & 18.2\% & 17 \\
		{\textbf{At}} & 14.40 & 65.71 & 16.0\% & 18 \\
		{\textbf{Fo}} & 14.92 & 63.77 & 17.4\% & 18 \\
		{\textbf{Tr}} & 13.97 & 59.77 & 18.1\% & 17 \\
		{\textbf{Sh}} & 13.12 & 55.53 & 16.4\% & 16 \\
		\midrule
		{\textbf{Sc}} & 11.07 & 52.88 & 21.3\% & 14 \\
		{\textbf{Na}} &  7.34 & 34.29 & 24.6\% & 10 \\
		{\textbf{We}} &  7.45 & 36.40 &  3.8\% & 8 \\
		\bottomrule
	\end{tabular}
	\caption{\footnotesize Overview of few-shot multi-intent detection data from TourSG (above midline) and StanfordLU (below midline). 
		$|S|$ is the average support set size. 
		P. ML denotes the proportion multi-label sentences. 
		$|\mathcal{Y}|$ is the label numbers. 
	}\label{tbl:new_dataset}
\end{table}

\paragraph{Few-shot Data Construction}
To simulate the few-shot situation, we reconstruct the dataset into few-shot learning form, where each sample is the combination of a query instance $(\bm{x^q},\bm{y^q})$ and corresponding K-shot support set $\mathcal{S}$.
Table \ref{tbl:new_dataset} shows the overview of the experiment data.

Different from the single-label classification problem, multi-label instance is associated with multiple labels.
There, we cannot guarantee that each label appears $K$ times while sampling the support sentences.
To cope with this, we approximately construct K-shot support set $\mathcal{S}$ with the Minimum-including Algorithm \cite{hou2020fewshot}. 
It constructs support set generally following two criteria:  
(1) All labels within the domain appear at least $K$ times in $\mathcal{S}$.  
(2) At least one label will appear less than $K$ times in $\mathcal{S}$ if any $(\bm{x},\bm{y})$ pair is removed from it.\footnote{
	Removing steps have a preference for instances with more labels. 
	So we randomly skip removing by the chance of 20\%. 
}

For each domain, we sample $N_s$ different $K$-shot support sets. 
Then, for each support set, we sample $N_q$ unincluded utterances as queries (query set). 
Each support-query-set pair forms one \textbf{few-shot episode}. 
Eventually, we get $N_s$ episodes and $N_s \times N_q$ samples for each domain.

For TourSG, we construct 100 few-shot episodes for each source domain and 50 few-shot episodes for each target domain. 
And the query set size is 16. 
Because StanfordLU has fewer domains, we construct 200 few-shot episodes for each source domain and 50 few-shot episodes for each target domain. 
And query set size is 32.  


\paragraph{Evaluation}
To conduct robust evaluation under few-shot setting, 
we cross-validate the models on different domains.
Each time,
we pick one target domain for testing, one domain for development, 
and use the rest domains of the same dataset as source domains for training.\footnote{
For example of the TourSG dataset, each round, model is trained on $4 \times 100 \times 16 = 6400$ samples, 
validated on $1 \times 50 \times 16 = 1600$ samples, and tested on $1 \times 50 \times 16 = 800$ samples. 
Then, with all the 6 cross-evaluation rounds, each model is totally tested on $800 \times 6 = 4800$ samples. }

When testing model on a target domain, 
we evaluate micro F1 scores within each few-shot episode.
Then we average F1 scores from all episodes as the final result to counter the randomness from support-sets.
To control the nondeterminacy of neural network training \citep{reimers-gurevych:2017:EMNLP2017}, 
we report the average score of 5 random seeds.

\paragraph{Implements}
For sentence embedding and label name, we average the token embedding provided by pretrained language model and we use  
\textit{Electra-small} \cite{clark2020electra} and uncased \textit{BERT-Base} \citep{BERT} here.
Besides, we adopt embedding tricks of Pairs-Wise Embedding \cite{hou2020fewshot} and Gradual Unfreezing \cite{Howard2018UniversalLM}.
We use ADAM \citep{DBLP:journals/corr/KingmaB14} to train the models with batch size 4. 
Learning rate is set as 1e-5 for both our model and baseline models. 
We set $\alpha$ (Eq. \ref{eq:alpha}) as 0.3 and vary $\beta$ (Eq. \ref{eq:beta}) in \{0.1, 0.5, 0.9\} 
considering label name's anchoring power with different datasets and support-set sizes. 
For the MLP of kernel regression, we employ ReLU as activation function and vary the layers in \{1, 2, 3\} and hidden dimension in \{5, 10, 20\}. 
The best hyperparameter are determined on the development domains.

\subsection{Baselines}
We compare our model with two kinds of strong baseline: 
fine-tune based transfer learning methods (TransferM) and similarity-based FSL methods (MPN and MMN).

\textbf{TransferM} is a domain transfer model with large pretrained language model and a multi-label classification layer. 
Following popular MLC settings, we use a fixed threshold tuned on dev set. 
We pretrain it on source domains and select the best model on the dev set. 
We deal with mismatch of label set by re-training classifying layers for different domains. 
On target domain, model is fine-tuned on support set. 

\textbf{Multi-label Prototypical Network (MPN)} is a similarity based few-shot learning model 
that calculates sentence-label relevance score with a prototypical network \citep{prototypical} and 
uses a fixed threshold tuned on dev set. 
It is pre-trained on source domains and directly works on target domains without fine-tuning. 

\textbf{Multi-label Matching Network (MMN)} is all the same as MPN but employs the Matching Network \cite{matching} for label-instance relevance score calculation.

\subsection{Main Results}\label{sec:main_res}
Here we evaluate the proposed method on both 1-shot and 5-shot multi-label intent detection. 

\begin{table*}[t]
	\centering
	\footnotesize
		\begin{tabular}{llrrrrrrrrrrr}\toprule
			& 
			\multirow{2}{*}{\textbf{Model}} &
			\multicolumn{6}{c}{\textbf{TourSG}} & &
			\multicolumn{3}{c}{\textbf{StanfordLU}}	& \\
			\cmidrule(lr){3-8} 
			\cmidrule(lr){10-12}
			& & \multicolumn{1}{c}{\textbf{It}} & \multicolumn{1}{c}{\textbf{Ac}} & \multicolumn{1}{c}{\textbf{At}} & \multicolumn{1}{c}{\textbf{Fo}} & \multicolumn{1}{c}{\textbf{Tr}} & \multicolumn{1}{c}{\textbf{Sh}} & \multicolumn{1}{c}{\textbf{Ave.}} &
			\multicolumn{1}{c}{\textbf{Sc}} & \multicolumn{1}{c}{\textbf{Na}} & \multicolumn{1}{c}{\textbf{We}} & \multicolumn{1}{c}{\textbf{Ave.}} \\
			\midrule
			\multirow{5}{*}{\textbf{+E}}
			& TransferM & { 14.34 } & { 14.75 } & { 16.13 } & { 11.79 } & { 13.64 } & { 14.32 } & { 14.16 } & { 16.96 } & { 22.99 } & { 21.01 } & { 20.32 }\\
			& MMN & { 9.98 } & { 7.81 } & { 8.37 } & { 7.81 } & { 10.65 } & { 11.56 } & { 9.36 } & { 31.22 } & { 24.41 } & { \textbf{48.01} } & { 34.55 }\\
			& MPN & { 12.24 } & { 10.38 } & { 10.00 } & { 10.47 } & { 13.61 } & { 11.41 } & { 11.35 } & { 32.44 } & { 17.83 } & { 38.86 } & { 29.71 }\\
			& MPN+ALR & { 28.74 } & { 34.94 } & { 35.06 } & { 34.62 } & { 35.53 } & { 31.87 } & { 33.46 } & { 33.35 } & { 28.88 } & { 45.58 } & { 35.93 }\\
			& Ours & { \textbf{39.98} } & { \textbf{51.55} } & { \textbf{55.16} } & { \textbf{52.16} } & { \textbf{55.36} } & { \textbf{52.20} } & { \textbf{51.07} } & { \textbf{40.61} } & { \textbf{40.76} } & { 46.16 } & { \textbf{42.51} }\\
			\midrule
			\multirow{5}{*}{\textbf{+B}}
			& TransferM & { 16.78 } & { 18.62 } & { 14.92 } & { 16.40 } & { 15.68 } & { 14.50 } & { 16.15 } & { 18.00 } & { 24.65 } & { 22.26 } & { 21.64 }\\
			& MMN & { 10.89 } & { 7.72 } & { 8.92 } & { 9.32 } & { 13.75 } & { 10.87 } & { 10.24 } & { 39.18 } & { 35.35 } & { 45.87 } & { 40.13 }\\
			& MPN & { 13.77 } & { 12.38 } & { 13.46 } & { 10.23 } & { 16.19 } & { 15.79 } & { 13.64 } & { 39.34 } & { 36.09 } & { 45.86 } & { 40.43 }\\
			& MPN+ALR & { 40.99 } & { 51.57 } & { 54.91 } & { 51.90 } & { 54.87 } & { 50.76 } & { 50.83 } & { 38.81 } & { 41.08 } & { \textbf{54.16} } & { 44.68 }\\
			& Ours & { \textbf{44.58} } & { \textbf{57.11} } & { \textbf{60.34} } & { \textbf{56.49} } & { \textbf{60.18} } & { \textbf{55.60} } & { \textbf{55.72} } & { \textbf{42.55} } & { \textbf{56.95} } & { 53.14 } & { \textbf{50.88} }\\
			\bottomrule
		\end{tabular}
	\caption{ \footnotesize  
		F1 scores on 1-shot multi-label intent detection. 
		\textbf{+E} and \textbf{+B} denote use Electra-small (14M params) and BERT-base (110M params) as embedder respectively. 
		\textbf{Ave.} shows the averaged scores. Results with standard deviations are included in Appendix.
	}\label{tbl:1shot}
\end{table*}

\begin{table*}[t]
	\centering
	\footnotesize
		\begin{tabular}{llrrrrrrrrrrr}\toprule
			& 
			\multirow{2}{*}{\textbf{Model}} &
			\multicolumn{6}{c}{\textbf{TourSG}} & &
			\multicolumn{3}{c}{\textbf{StanfordLU}}	& \\
			\cmidrule(lr){3-8} 
			\cmidrule(lr){10-12}
			& & \multicolumn{1}{c}{\textbf{It}} & \multicolumn{1}{c}{\textbf{Ac}} & \multicolumn{1}{c}{\textbf{At}} & \multicolumn{1}{c}{\textbf{Fo}} & \multicolumn{1}{c}{\textbf{Tr}} & \multicolumn{1}{c}{\textbf{Sh}} & \multicolumn{1}{c}{\textbf{Ave.}} &
			\multicolumn{1}{c}{\textbf{Sc}} & \multicolumn{1}{c}{\textbf{Na}} & \multicolumn{1}{c}{\textbf{We}} & \multicolumn{1}{c}{\textbf{Ave.}} \\
			\midrule
			\multirow{5}{*}{\textbf{+E}}
			 & TransferM & { 14.72 } & { 19.20 } & { 16.18 } & { 18.86 } & { 17.17 } & { 17.51 } & { 17.27 } & { 16.99 } & { 23.79 } & { 23.92 } & { 21.57 }\\
			 & MMN & { 14.11 } & { 10.58 } & { 17.80 } & { 12.74 } & { 18.01 } & { 16.76 } & { 15.00 } & { 41.91 } & { 37.94 } & { \textbf{60.67} } & { 46.84 }\\
			 & MPN & { 15.18 } & { 15.56 } & { 17.60 } & { 15.01 } & { 17.99 } & { 17.17 } & { 16.42 } & { 35.92 } & { 27.65 } & { 58.07 } & { 40.55 }\\
			 & MPN+ALR & { 29.74 } & { 30.91 } & { 34.28 } & { 33.61 } & { 35.90 } & { 33.44 } & { 32.98 } & { 44.52 } & { 42.39 } & { 54.42 } & { 47.11 }\\
			 & Ours & { \textbf{44.21} } & { \textbf{51.37} } & { \textbf{55.76} } & { \textbf{54.50} } & { \textbf{55.37} } & { \textbf{54.55} } & { \textbf{52.63} } & { \textbf{51.83} } & { \textbf{46.44} } & { 54.17 } & { \textbf{50.82} }\\
			\midrule
			\multirow{5}{*}{\textbf{+B}}
			 & TransferM & { 17.98 } & { 16.51 } & { 19.88 } & { 17.22 } & { 13.84 } & { 15.41 } & { 16.81 } & { 16.62 } & { 23.69 } & { 26.64 } & { 22.31 }\\
			 & MMN & { 15.65 } & { 16.42 } & { 19.90 } & { 12.23 } & { 16.81 } & { 17.13 } & { 16.36 } & { 43.65 } & { 51.94 } & { 46.65 } & { 47.41 }\\
			 & MPN & { 20.71 } & { 22.39 } & { 26.51 } & { 21.94 } & { 23.41 } & { 24.52 } & { 23.24 } & { 41.45 } & { 50.51 } & { 54.96 } & { 48.97 }\\
			 & MPN+ALR & { 45.51 } & { 53.71 } & { 58.16 } & { 56.91 } & { 57.62 } & { 54.86 } & { 54.46 } & { 51.30 } & { 47.80 } & { \textbf{60.08} } & { 53.06 }\\
			 & Ours & { \textbf{46.80} } & { \textbf{54.79} } & { \textbf{59.95} } & { \textbf{59.11} } & { \textbf{60.13} } & { \textbf{58.56} } & { \textbf{56.56} } & { \textbf{52.17} } & { \textbf{60.36} } & { 59.63 } & { \textbf{57.39} }\\
			\bottomrule
		\end{tabular}
	\caption{ \footnotesize  
		F1 score results on 5-shot multi-label intent detection. 
		Standard deviations are included in Appendix.
	}\label{tbl:5shot}
\end{table*}

\paragraph{Result of 1-shot setting.}
Table \ref{tbl:1shot} shows the results of 1-shot multi-label intent detection. 
Each column respectively shows the F1 scores of taking a certain domain as target domain (test) and use others as source domain (train \& dev).
When using BERT embedding, our model outperforms the strongest baseline by average F1 scores of 39.57 on TourSG and 10.45 on Stanford dataset.
Our improvement on TourSG is much higher than those on StanfordLU.
We find the gap mainly comes from the difference in label set characteristics of two datasets and will analyze this latter.

BERT is sometimes too computational costly in real-world applications. 
So we also evaluate model performance with lighter embedding of Electra-small. 
It only has 14M parameters, which is much smaller than the 110M of BERT. 
Again, our model achieves the best performance with Electra. 
Interestingly, our Electra-based model is even better than all baselines using BERT, 
which is especially valuable in scenarios with limited computing resources.

When comparing to traditional transferring-based method (TransferM), 
all non-finetune-based methods (Ours, MPN and MMN) gain huge improvements on StanfordLU dataset. 
This mainly comes from the superiority of the non-finetune-based in overfitting resistance. 
For TourSG, domain gaps are lower and labels of different domains are similar, which makes it easier to transfer source domain parameters.  
As a result, TransferM performs slightly better than MPN on TourSG.
In contrast, our models have stable performance on both types of datasets, which reflects the model versatility.

Our model can be regarded as \textit{MPN+ALR+MCT}. Thus, the stepped growth between \textit{MPN}, \textit{MPN+ALR} and \textit{Ours} demonstrates the effectiveness of Both ALR and MCT.

\paragraph{Result of 5-shot setting.}
Table \ref{tbl:5shot} shows the results of 5-shots experiments.
The results are consistent with 1-shot setting in general trending. 
Our methods achieve the best performance. 
By comparing 1-shot and 5-shots results, we find that our 1-shot model is able to outperform most 5-shot baselines.
This indicates that our model can better exploit prior experience and rely less on support examples.

\subsection{Analysis}\label{sec:ana}

\paragraph{Ablation Test}

%

\begin{table}[t]
	\centering
	\footnotesize
		\begin{tabular}{lrrrr}
			\toprule
			\multirow{2}{*}{\textbf{Setting}} & \multicolumn{2}{c}{\textbf{TourSG}} & \multicolumn{2}{c}{\textbf{StanfordLU}}  \\
			\cmidrule(lr){2-3}
			\cmidrule(lr){4-5}
			& \multicolumn{1}{c}{1-shot} & \multicolumn{1}{c}{5-shots} & \multicolumn{1}{c}{1-shot} & \multicolumn{1}{c}{5-shots} \\
			\midrule
			Ours & { 51.07 } & { 52.63 } & { 42.51 } & { 50.82 }\\
			~ - ALR  & { -38.53 } & { -31.33 } & { -11.33 } & { -10.31 }\\
			~ - MCT & { -12.52 } & { -16.70 } & { -10.12 } & { -17.45 }\\
			\bottomrule
		\end{tabular}
	\caption{
		\footnotesize
		Ablation study over two main components of proposed framework: Anchored Label Representation and Meta Calibrated Threshold. 
		Result is the averaged accuracy of all domains.
	}\label{tbl:ablation}
		\vspace*{-3mm}
	
\end{table}

\begin{table}[t]
	\centering
	\footnotesize
		\begin{tabular}{lrrrr}
			\toprule
			\multirow{2}{*}{\textbf{Setting}} & \multicolumn{2}{c}{\textbf{TourSG}} & \multicolumn{2}{c}{\textbf{StanfordLU}}  \\
			\cmidrule(lr){2-3}
			\cmidrule(lr){4-5}
			& \multicolumn{1}{c}{1-shot} & \multicolumn{1}{c}{5-shots} & \multicolumn{1}{c}{1-shot} & \multicolumn{1}{c}{5-shots} \\
			\midrule
			MCT (Ours) & { 51.07 } & { 52.63 } & { 42.51 } & { 50.82 }\\
			~ - Logits Adapting  & { -6.13 } & { -7.36 } & { -2.45 } & { -7.78 }\\
			~ - Meta Learning & { -2.73 } & { -3.36 } & { -3.59 } & { -9.35 }\\
			~ - KR Calibration & { -0.86 } & { -2.28 } & { -2.08 } & { -8.96 }\\
			\bottomrule
		\end{tabular}
	\caption{
		\footnotesize
		Detailed ablation test over Meta Calibrated Threshold.
	}\label{tbl:MCT_ablation}
	
\end{table}

To understand the contribution of each framework component, 
we conduct 1-shot/5-shots ablation study with Electra embedding in Table \ref{tbl:ablation}.
We independently remove two main components: \textit{Anchored Label Representation} (ALR) and \textit{Meta Calibrated Threshold} (MCT). 

When ALR is removed, we represent each label with only prototypical embeddings constructed from support examples. 
Huge F1 score drops are witnessed especially on TourSG.
On one hand, TourSG has similar labels across different domains, which greatly benefits the label embedding learning of ALR.
On the other hand, model without ALR are often confused by co-occurring intents, such as ``thank'' and ``confirm'', which can be easily separated by ALR.

For our model without MCT, 
we use a vanilla threshold tuned on source domains. 
We find MCT has more impacts on 5-shot settings.
This is because logits adapting and KR calibration of MCT exploit the relation between specific query and support set, which promotes model to benefits from more support examples.

By comparing the opposite influence of shot number on ALR and MCT, we found our framework reaches a balance, since its two components are respectively good at transferring prior knowledge and exploiting domain knowledge. 

\paragraph{Analysis over components of Meta Calibrate Threshold}
We further disassemble the MCT and to see the contributions of the three sub-components.

When we remove \textit{Logits-Adapting}, we use a single learnable value as meta-threshold. 
The performance drops indicate that Logits-Adaptiving threhold provides better domain generalization than traditional threshold values. 

If we remove \textit{Meta Learning} of threshold, we replace the meta threshold with a fixed Logits-Adapting threshold, and calibrate it without learning of meta parameters in Kernel Regression. 
We address the performance loss to the fact that meta learning process provides prior experience which effectively aids the non-parametric learning of target domains.

For our model without \textit{KR Calibration}, we directly predict labels with meta thresholds. 
The score drops show calibration helps by adapting thresholds to different domains. 
The drops on TourSG is limited, because the domain gap of TourSG is small, and meta threshold learned on source domains are often good enough even without calibration.

%
%

\begin{table}[t]
	\centering
	\footnotesize
	\begin{tabular}{lrrrr}
		\toprule
		\multirow{2}{*}{\textbf{Setting}} & \multicolumn{2}{c}{\textbf{TourSG}} & \multicolumn{2}{c}{\textbf{StanfordLU}}  \\
		\cmidrule(lr){2-3}
		\cmidrule(lr){4-5}
		& \multicolumn{1}{c}{1-shot} & \multicolumn{1}{c}{5-shots} & \multicolumn{1}{c}{1-shot} & \multicolumn{1}{c}{5-shots} \\
		\midrule
		ALR   & 68.16 & 67.28 & 15.67 & 21.91 \\
		ALR + MT  & 77.85 & 78.18 & 51.24 & 51.38 \\
		ALR + MT + KR  & \textbf{82.26} & \textbf{82.05} & \textbf{80.92} & \textbf{84.70} \\
		\bottomrule
	\end{tabular}
	\caption{
		\footnotesize
		Analysis of label number accuracy. 
		ALR denotes model with Anchored Label Representation. 
		MT is Meta Threshold.
		KR is Calibration with Kernel-Regression. 
	}\label{tbl:label_acc}
	\vspace*{-3mm}
\end{table}

\paragraph{Label Number Accuracy Analysis}
To understand the impact of thresholding module, 
we conduct accuracy analysis of whether model can predict correct number of labels. 
As Table \ref{tbl:label_acc} presents,  
when adding Meta Threshold and KR Calibration, we can observe a continuous increase in the label number accuracy.
This shows that both Meta Threshold and KR Calibration can greatly help model to decide proper label numbers.

\section{Related Work}

Usually, multi-label classification (MLC) methods rely on thresholds to predict multiple labels. 
For MLC problem in NLP,
\citet{wu2019learning} leverage meta learning to estimate thresholds for data-rich setting. 
For thresholding of intent detection, \citet{gangadharaiah2019joint} leverage a fixed threshold over intent possibility. 
\citet{xu2017convolutional} learn threshold with linear regression.

Without threshold, one solution to MLC is Label Powerset (LP) \cite{tsoumakas2010random,tsoumakas2007random}, which regards combination of multiple labels as a single label. \citet{xu2013exploiting} explore idea of LP in multi-label intent detection. However, LP often suffers from data sparseness from label combination even in data-rich settings.
In addition to LP, \citet{kim2017two} propose to learn multi-intent detection from single intent data. 
They first detect sub-sentence, and then predict single intents on sub-sentences. 
But, their method is limited by the explicit conjunctions in data, and it is hard to learn to detect sub-sentence in few-shot setting. 

Few-shot learning in NLP has been widely explored for single-label classification tasks, including text classification \cite{textSun2019hierarchical,textGeng2019induction,yan2018few,yu2018diverse,DBLP:conf/iclr/BaoWCB20,policyVlasov2018few}, relation classification \cite{relationLv2019adapting,relationGao2019neural,relationYeL19}, sequence labeling \cite{hou2020fewshot}.
However, few-shot multi-label problem is less investigated. 
Previous works focus on computer vision \cite{xiang2019incremental,alfassy2019laso} and signal processing \cite{cheng2019multi}.
\citet{rios2018few} investigate few-shot MLC for medical texts. 
But, their method requires descriptions and EMR structure of labels, which are often hard to obtain and not available in our task. 
For the use of label name semantics, it has been proven to be effective for data scarcity problem of both slot filling \cite{zeroBapna2017towards,zeroLee2019zero,zeroSlotShah2019robust,hou2020fewshot} and intent detection \cite{xia2020cg,krone2020learning,chen2016zero}. 
Our method shares the similar idea but introduces it to tackle the special challenges of multi-label setting. 

\section{Conclusion}

In this paper, we explore the few-shot learning problem of multi-label intent detection. 
To estimate a reasonable threshold with only a few support examples, 
we propose the Meta Calibrated Threshold that adaptively combines prior experience and domain-specific knowledge. 
To obtain label-instance relevance score under few-shot setting, 
we introduce a metric learning based method with Anchored Label Representation.
It provides well-separated label representations for label-instance similarity calculation. 
Experiment results validate that both the Meta Calibrated Threshold and Anchored Label Representation can improve the few-shot multi-label intent detection. 

\bibliography{AAAI21_MLFS}

\appendix
\section*{Appendices}

\section{Full Results with Standard Deviations}

Here, we present the results with standard deviations for few-shot multi-label intent detection in Table \ref{tbl:1shot_toursg}, \ref{tbl:5shot_toursg}, \ref{tbl:1shot_stanford} and \ref{tbl:5shot_stanford}.

\begin{table*}[t]
	\centering
	\footnotesize
	\begin{tabular}{llrrrrrrrrrrr}\toprule
		&
		\multirow{1}{*}{\textbf{Model}} 
		& \multicolumn{1}{c}{\textbf{It}} & \multicolumn{1}{c}{\textbf{Ac}} & \multicolumn{1}{c}{\textbf{At}} & \multicolumn{1}{c}{\textbf{Fo}} & \multicolumn{1}{c}{\textbf{Tr}} & \multicolumn{1}{c}{\textbf{Sh}} & \multicolumn{1}{c}{\textbf{Ave.}} 
		
		\\
		\midrule
		\multirow{5}{*}{\textbf{+E}}
		& TransferM & { 14.34\tiny{$\pm 0.82$} } & { 14.75\tiny{$\pm 0.91$} } & { 16.13\tiny{$\pm 1.35$} } & { 11.79\tiny{$\pm 1.54$} } & { 13.64\tiny{$\pm 0.33$} } & { 14.32\tiny{$\pm 1.17$} } & { 14.16\tiny{$\pm 1.02$} }\\
		& MN & { 9.98\tiny{$\pm 1.80$} } & { 7.81\tiny{$\pm 0.70$} } & { 8.37\tiny{$\pm 0.86$} } & { 7.81\tiny{$\pm 0.35$} } & { 10.65\tiny{$\pm 1.62$} } & { 11.56\tiny{$\pm 0.79$} } & { 9.36\tiny{$\pm 1.02$} }\\
		& MPN & { 12.24\tiny{$\pm 0.92$} } & { 10.38\tiny{$\pm 1.21$} } & { 10.00\tiny{$\pm 0.54$} } & { 10.47\tiny{$\pm 0.42$} } & { 13.61\tiny{$\pm 0.92$} } & { 11.41\tiny{$\pm 0.31$} } & { 11.35\tiny{$\pm 0.72$} }\\
		& MPN+ALR & { 28.74\tiny{$\pm 2.18$} } & { 34.94\tiny{$\pm 1.91$} } & { 35.06\tiny{$\pm 3.83$} } & { 34.62\tiny{$\pm 2.69$} } & { 35.53\tiny{$\pm 1.97$} } & { 31.87\tiny{$\pm 2.31$} } & { 33.46\tiny{$\pm 2.48$} }\\
		& Ours & { \textbf{39.98}\tiny{$\pm 0.56$} } & { \textbf{51.55}\tiny{$\pm 1.53$} } & { \textbf{55.16}\tiny{$\pm 2.43$} } & { \textbf{52.16}\tiny{$\pm 0.98$} } & { \textbf{55.36}\tiny{$\pm 0.96$} } & { \textbf{52.20}\tiny{$\pm 1.03$} } & { \textbf{51.07}\tiny{$\pm 1.24$} }\\
		\midrule
		\multirow{5}{*}{\textbf{+B}}
		& TransferM & { 16.78\tiny{$\pm 0.05$} } & { 18.62\tiny{$\pm 0.59$} } & { 14.92\tiny{$\pm 2.22$} } & { 16.40\tiny{$\pm 2.58$} } & { 15.68\tiny{$\pm 0.32$} } & { 14.50\tiny{$\pm 2.18$} } & { 16.15\tiny{$\pm 1.32$} }\\
		& MMN & { 10.89\tiny{$\pm 3.35$} } & { 7.72\tiny{$\pm 1.44$} } & { 8.92\tiny{$\pm 1.45$} } & { 9.32\tiny{$\pm 1.40$} } & { 13.75\tiny{$\pm 0.70$} } & { 10.87\tiny{$\pm 4.31$} } & { 10.24\tiny{$\pm 2.11$} }\\
		& MPN & { 13.77\tiny{$\pm 0.38$} } & { 12.38\tiny{$\pm 0.32$} } & { 13.46\tiny{$\pm 0.14$} } & { 10.23\tiny{$\pm 0.30$} } & { 16.19\tiny{$\pm 0.19$} } & { 15.79\tiny{$\pm 0.38$} } & { 13.64\tiny{$\pm 0.28$} }\\
		& MPN+ALR & { 40.99\tiny{$\pm 1.54$} } & { 51.57\tiny{$\pm 1.04$} } & { 54.91\tiny{$\pm 0.31$} } & { 51.90\tiny{$\pm 1.98$} } & { 54.87\tiny{$\pm 0.82$} } & { 50.76\tiny{$\pm 1.30$} } & { 50.83\tiny{$\pm 1.17$} }\\
		& Ours & { \textbf{44.58}\tiny{$\pm 0.71$} } & { \textbf{57.11}\tiny{$\pm 1.22$} } & { \textbf{60.34}\tiny{$\pm 0.92$} } & { \textbf{56.49}\tiny{$\pm 0.67$} } & { \textbf{60.18}\tiny{$\pm 0.85$} } & { \textbf{55.60}\tiny{$\pm 0.66$} } & { \textbf{55.72}\tiny{$\pm 1.03$} }\\
		
		\bottomrule
	\end{tabular}
	\caption{ \footnotesize  
		F1 scores on 1-shot multi-label intent detection on TourSG dataset. 
		\textbf{+E} and \textbf{+B} denote use Electra-small and BERT-base as embedder respectively. 
		Ave. shows the averaged scores.
	}\label{tbl:1shot_toursg}
\end{table*}

\begin{table*}[t]
	\centering
	\footnotesize
	\begin{tabular}{llrrrrrrrrrrr}\toprule
		&
		\multirow{1}{*}{\textbf{Model}} 
		& \multicolumn{1}{c}{\textbf{It}} & \multicolumn{1}{c}{\textbf{Ac}} & \multicolumn{1}{c}{\textbf{At}} & \multicolumn{1}{c}{\textbf{Fo}} & \multicolumn{1}{c}{\textbf{Tr}} & \multicolumn{1}{c}{\textbf{Sh}} & \multicolumn{1}{c}{\textbf{Ave.}} 
		
		\\
		\midrule
		\multirow{5}{*}{\textbf{+E}}
		& TransferM & { 14.72\tiny{$\pm 0.53$} } & { 19.20\tiny{$\pm 1.59$} } & { 16.18\tiny{$\pm 1.03$} } & { 18.86\tiny{$\pm 1.04$} } & { 17.17\tiny{$\pm 1.19$} } & { 17.51\tiny{$\pm 1.63$} } & { 17.27\tiny{$\pm 1.17$} }\\
		& MN & { 14.11\tiny{$\pm 0.83$} } & { 10.58\tiny{$\pm 1.35$} } & { 17.80\tiny{$\pm 1.12$} } & { 12.74\tiny{$\pm 0.87$} } & { 18.01\tiny{$\pm 0.90$} } & { 16.76\tiny{$\pm 0.92$} } & { 15.00\tiny{$\pm 1.00$} }\\
		& MPN & { 15.18\tiny{$\pm 0.63$} } & { 15.56\tiny{$\pm 0.54$} } & { 17.60\tiny{$\pm 1.15$} } & { 15.01\tiny{$\pm 0.19$} } & { 17.99\tiny{$\pm 0.36$} } & { 17.17\tiny{$\pm 1.09$} } & { 16.42\tiny{$\pm 0.66$} }\\
		& MPN+ALR & { 29.74\tiny{$\pm 3.18$} } & { 30.91\tiny{$\pm 2.51$} } & { 34.28\tiny{$\pm 3.06$} } & { 33.61\tiny{$\pm 2.70$} } & { 35.90\tiny{$\pm 2.85$} } & { 33.44\tiny{$\pm 3.58$} } & { 32.98\tiny{$\pm 2.98$} }\\
		& Ours & { \textbf{44.21}\tiny{$\pm 0.71$} } & { \textbf{51.37}\tiny{$\pm 1.22$} } & { \textbf{55.76}\tiny{$\pm 0.92$} } & { \textbf{54.50}\tiny{$\pm 0.58$} } & { \textbf{55.37}\tiny{$\pm 0.95$} } & { \textbf{54.55}\tiny{$\pm 0.86$} } & { \textbf{52.63}\tiny{$\pm 0.87$} }\\
		\midrule
		\multirow{5}{*}{\textbf{+B}}
		& TransferM & { 17.98\tiny{$\pm 1.80$} } & { 16.51\tiny{$\pm 1.95$} } & { 19.88\tiny{$\pm 4.17$} } & { 17.22\tiny{$\pm 3.01$} } & { 13.84\tiny{$\pm 1.40$} } & { 15.41\tiny{$\pm 2.81$} } & { 16.81\tiny{$\pm 2.52$} }\\
		& MMN & { 15.65\tiny{$\pm 1.24$} } & { 16.42\tiny{$\pm 0.71$} } & { 19.90\tiny{$\pm 0.51$} } & { 12.23\tiny{$\pm 0.33$} } & { 16.81\tiny{$\pm 4.64$} } & { 17.13\tiny{$\pm 0.20$} } & { 16.36\tiny{$\pm 1.27$} }\\
		& MPN & { 20.71\tiny{$\pm 0.98$} } & { 22.39\tiny{$\pm 1.95$} } & { 26.51\tiny{$\pm 0.72$} } & { 21.94\tiny{$\pm 1.59$} } & { 23.41\tiny{$\pm 1.31$} } & { 24.52\tiny{$\pm 3.31$} } & { 23.24\tiny{$\pm 1.64$} }\\
		& MPN+ALR & { 45.51\tiny{$\pm 0.51$} } & { 53.71\tiny{$\pm 0.95$} } & { 58.16\tiny{$\pm 0.53$} } & { 56.91\tiny{$\pm 0.51$} } & { 57.62\tiny{$\pm 0.70$} } & { 54.86\tiny{$\pm 0.59$} } & { 54.46\tiny{$\pm 0.63$} }\\
		& Ours & { \textbf{46.80}\tiny{$\pm 0.83$} } & { \textbf{54.79}\tiny{$\pm 0.80$} } & { \textbf{59.95}\tiny{$\pm 0.46$} } & { \textbf{59.11}\tiny{$\pm 0.39$} } & { \textbf{60.13}\tiny{$\pm 0.44$} } & { \textbf{58.56}\tiny{$\pm 0.30$} } & { \textbf{56.56}\tiny{$\pm 0.54$} }\\
		
		\bottomrule
	\end{tabular}
	\caption{ \footnotesize  
		F1 scores on 5-shot multi-label intent detection on TourSG dataset. 
		Ave. shows the averaged scores. 
	}\label{tbl:5shot_toursg}
\end{table*}

\begin{table*}[t]
	\centering
	\footnotesize
	\begin{tabular}{llrrrrrrrrrrr}\toprule
		&
		\multirow{1}{*}{\textbf{Model}} 
		& \multicolumn{1}{c}{\textbf{Sc}} & \multicolumn{1}{c}{\textbf{Na}} & \multicolumn{1}{c}{\textbf{We}} & \multicolumn{1}{c}{\textbf{Ave.}} 
		
		\\
		\midrule
		\multirow{5}{*}{\textbf{+E}}
		& TransferM & { 16.96\tiny{$\pm 0.73$} } & { 22.99\tiny{$\pm 0.51$} } & { 21.01\tiny{$\pm 0.57$} } & { 20.32\tiny{$\pm 0.60$} }\\
		& MN & { 31.22\tiny{$\pm 4.96$} } & { 24.41\tiny{$\pm 3.28$} } & { \textbf{48.01}\tiny{$\pm 1.10$} } & { 34.55\tiny{$\pm 3.11$} }\\
		& MPN & { 32.44\tiny{$\pm 3.75$} } & { 17.83\tiny{$\pm 3.83$} } & { 38.86\tiny{$\pm 4.18$} } & { 29.71\tiny{$\pm 3.92$} }\\
		& MPN+ALR & { 33.35\tiny{$\pm 1.00$} } & { 28.88\tiny{$\pm 1.57$} } & { 45.58\tiny{$\pm 1.98$} } & { 35.93\tiny{$\pm 1.52$} }\\
		& Ours & { \textbf{40.61}\tiny{$\pm 1.05$} } & { \textbf{40.76}\tiny{$\pm 0.89$} } & { 46.16\tiny{$\pm 0.96$} } & { \textbf{42.51}\tiny{$\pm 0.97$} }\\
		\midrule
		\multirow{5}{*}{\textbf{+B}}
		& TransferM & { 18.00\tiny{$\pm 0.62$} } & { 24.65\tiny{$\pm 0.79$} } & { 22.26\tiny{$\pm 0.64$} } & { 21.64\tiny{$\pm 0.68$} }\\
		& MMN & { 39.18\tiny{$\pm 0.52$} } & { 35.35\tiny{$\pm 1.72$} } & { 45.87\tiny{$\pm 2.81$} } & { 40.13\tiny{$\pm 1.68$} }\\
		& MPN & { 39.34\tiny{$\pm 1.38$} } & { 36.09\tiny{$\pm 0.77$} } & { 45.86\tiny{$\pm 2.50$} } & { 40.43\tiny{$\pm 1.55$} }\\
		& MPN+ALR & { 38.81\tiny{$\pm 1.13$} } & { 41.08\tiny{$\pm 1.76$} } & { \textbf{54.16}\tiny{$\pm 2.12$} } & { 44.68\tiny{$\pm 1.67$} }\\
		& Ours & { \textbf{42.55}\tiny{$\pm 0.40$} } & { \textbf{56.95}\tiny{$\pm 0.77$} } & { 53.14\tiny{$\pm 1.89$} } & { \textbf{50.88}\tiny{$\pm 1.02$} }\\
		\bottomrule
	\end{tabular}
	\caption{ \footnotesize  
		F1 scores on 1-shot multi-label intent detection on StanfordLU dataset. 
		\textbf{+E} and \textbf{+B} denote use Electra-small and BERT-base as embedder respectively. 
		Ave. shows the averaged scores. 
	}\label{tbl:1shot_stanford}
\end{table*}

\begin{table*}[t]
	\centering
	\footnotesize
	\begin{tabular}{llrrrrrrrrrrr}\toprule
		&
		\multirow{1}{*}{\textbf{Model}} 
		& \multicolumn{1}{c}{\textbf{Sc}} & \multicolumn{1}{c}{\textbf{Na}} & \multicolumn{1}{c}{\textbf{We}} & \multicolumn{1}{c}{\textbf{Ave.}} 
		
		\\
		\midrule
		\multirow{5}{*}{\textbf{+E}}
		& TransferM & { 16.99\tiny{$\pm 0.94$} } & { 23.79\tiny{$\pm 0.27$} } & { 23.92\tiny{$\pm 1.78$} } & { 21.57\tiny{$\pm 1.00$} }\\
		& MN & { 41.91\tiny{$\pm 4.49$} } & { 37.94\tiny{$\pm 1.38$} } & { \textbf{60.67}\tiny{$\pm 1.23$} } & { 46.84\tiny{$\pm 2.37$} }\\
		& MPN & { 35.92\tiny{$\pm 2.79$} } & { 27.65\tiny{$\pm 4.58$} } & { 58.07\tiny{$\pm 1.88$} } & { 40.55\tiny{$\pm 3.08$} }\\
		& MPN+ALR & { 44.52\tiny{$\pm 6.21$} } & { 42.39\tiny{$\pm 2.32$} } & { 54.42\tiny{$\pm 4.78$} } & { 47.11\tiny{$\pm 4.44$} }\\
		& Ours & { \textbf{51.83}\tiny{$\pm 1.31$} } & { \textbf{46.44}\tiny{$\pm 1.60$} } & { 54.17\tiny{$\pm 1.70$} } & { \textbf{50.82}\tiny{$\pm 1.54$} }\\
		\midrule
		\multirow{5}{*}{\textbf{+B}}
		& TransferM & { 16.62\tiny{$\pm 0.18$} } & { 23.69\tiny{$\pm 0.46$} } & { 26.64\tiny{$\pm 2.04$} } & { 22.31\tiny{$\pm 0.89$} }\\
		& MMN & { 43.65\tiny{$\pm 6.24$} } & { 51.94\tiny{$\pm 1.03$} } & { 46.65\tiny{$\pm 0.48$} } & { 47.41\tiny{$\pm 2.58$} }\\
		& MPN & { 41.45\tiny{$\pm 2.83$} } & { 50.51\tiny{$\pm 2.94$} } & { 54.96\tiny{$\pm 9.76$} } & { 48.97\tiny{$\pm 5.18$} }\\
		& MPN+ALR & { 51.30\tiny{$\pm 1.69$} } & { 47.80\tiny{$\pm 3.73$} } & { \textbf{60.08}\tiny{$\pm 2.64$} } & { 53.06\tiny{$\pm 2.69$} }\\
		& Ours & { \textbf{52.17}\tiny{$\pm 1.29$} } & { \textbf{60.36}\tiny{$\pm 1.55$} } & { 59.63\tiny{$\pm 2.23$} } & { \textbf{57.39}\tiny{$\pm 1.69$} }\\
		
		\bottomrule
	\end{tabular}
	\caption{ \footnotesize  
		F1 scores on 5-shot multi-label intent detection on StanfordLU dataset. 
		Ave. shows the averaged scores. 
	}\label{tbl:5shot_stanford}
\end{table*}

\end{document}